\documentclass{article}
\usepackage{authblk}
\usepackage{hyperref}
\usepackage{xcolor}
\usepackage{amsmath}
\usepackage{enumitem}
\usepackage{float}
\usepackage{caption}
\usepackage{subcaption}
\usepackage[style=apa]{biblatex} 
\bibliography{main}

\usepackage{tikz}

\usepackage{tikz-network}
\usepackage{fancyhdr}

\title{Qualities, challenges and future of genetic algorithms: a literature review\\
\textcolor{blue}{Early draft, feedback is welcome}}
\author[1,2,3]{Aymeric Vi\'{e}}
\author[1,2,4]{Alissa M. Kleinnijenhuis}
\author[1,2]{Doyne J. Farmer}
\affil[1]{Mathematical Institute, University of Oxford, UK}
\affil[2]{Institute of New Economic Thinking, University of Oxford, UK}
\affil[3]{Cross Labs, Cross Compass Ltd., Tokyo, Japan}
\affil[4]{Stanford Institute for Economic Policy Research, Stanford University}

\date{\today}
\begin{document}
\maketitle

\begin{abstract}
Genetic algorithms, computer programs that simulate natural evolution, are increasingly applied across many disciplines. 
They have been used to solve various optimisation problems from neural network architecture search to strategic games, and to model phenomena of adaptation and learning. 
Expertise on the qualities and drawbacks of this technique is largely scattered across the literature or former, motivating an compilation of this knowledge at the light of the most recent developments of the field.
In this review, we present genetic algorithms, their qualities, limitations and challenges, as well as some future development perspectives. 
Genetic algorithms are capable of exploring large and complex spaces of possible solutions, to quickly locate promising elements, and provide an adequate modelling tool to describe evolutionary systems, from games to economies. They however suffer from high computation costs, difficult parameter configuration, and crucial representation of the solutions. Recent developments such as GPU computing, parallel and quantum computing, conception of powerful parameter control methods, and novel approaches in representation strategies, may be keys to overcome those limitations. 
This compiling review aims at informing practitioners and newcomers in the field alike in their genetic algorithm research, and at outlining promising avenues for future research. It highlights the potential for interdisciplinary research associating genetic algorithms to pulse original discoveries in social sciences, open ended evolution, artificial life and AI.
    
\end{abstract}

\clearpage
\tableofcontents
\clearpage

\section{Introduction}

Inspired from Darwin's theory of evolution, the Genetic Algorithm (GA) is an adaptive search algorithm that simulates some of the evolution processes: selection, fitness, reproduction, crossover (also denoted recombination), mutation. A population of individuals (organisms, strategies, individuals, objects...) characterized by a genetic sequence inducing physical characteristics, evolves under selection pressure in an artificial environment. The fittest individuals reproduce and transmit their genes, while mutations and exchanges of genetic material explore new individual characteristics. \\


While originally designed for the study of adaptation in natural systems (\cite{holland1975adaptation}), most GAs have been developed for optimization purposes (\cite{whitley1994genetic}). They have been applied in a large variety of research domains: biology (\cite{street1999computational}), economics (\cite{chatterjee2018efficient}, \cite{waheeb2019new}), finance (\cite{lwin2014learning}, \cite{han2019using}), operational research (\cite{della1995genetic}, \cite{baker2003genetic}), game theory (\cite{axelrod1987evolution}, \cite{vie2020blotto}), deep learning and neural networks (\cite{stanley2019designing}, \cite{chung2020genetic}), forecasting (\cite{packard1988adaptation}, \cite{ahn2003elitism}), optimisation (\cite{Wirsansky2020hands}, \cite{dhunny2020economic}), computer science and algorithms (\cite{koza1992genetic}), healthcare (\cite{tao2019ga}, \cite{devarriya2020unbalanced}) or data science (\cite{yang1998feature}, \cite{maulik2000genetic}, \cite{raymer2000dimensionality}). \\

Knowledge on the qualities and drawbacks of this technique is largely scattered across the literature, dispersed over different fields, or former. Since the pioneering works of \cite{holland1975adaptation} or \cite{mitchell1994genetic},  applications of GAs have expanded and diversified. While several recent articles put an emphasis on the presentation of the GA technique, such as the operators being used (\cite{mirjalili2019genetic}) or its applications (\cite{kramer2017genetic}), an up to date compilation focused on the qualities and challenges is to the best of our knowledge, lacking and potentially of great interest to the field. \\

In this review, we introduce genetic algorithms, their key concepts and algorithm steps. We first discuss their qualities as a search method to understand the reasons behind their wide-ranging successfully applications, in both optimisation and modelling purposes. This discussion will then focus on the main limitations faced by GAs: their high computational cost, the difficulty of setting the GA parameters, and of adopting a proper representation of the solutions, have often been hurdles to their use. After highlighting these challenges, recent promising ways to overcome these constraints will be reviewed.\\


Describing their key qualities, we will first present the exploration properties of genetic algorithms (GAs). Their combination of parallel population-based search, with the support of the evolution operators, give to this search method a strong advantage in handling spaces of potential solutions that are large (\cite{axelrod1987evolution}), have local optima (\cite{maulik2000genetic}), admit several solutions (\cite{Wirsansky2020hands}), or have a complex structure challenging traditional optimisation tools. GAs achieve balance between exploration and exploitation, using previous evaluations to decide what to evaluate next. They quickly identify promising regions of that space with regards to an objective (\cite{de1993genetic}), and eventually locate the global optima (\cite{bhandari1996genetic}, \cite{del2001asymptotic}), making them relevant to solve difficult problems. GAs thrive in contexts where information and solution structure are scarce, and offer high flexibility. As robust, "weak" methods, GAs work well to approach phenomena or problems with unknown structure (\cite{whitley1994genetic}). They have solved high dimensional problems unsolvable by traditional reward-maximizing algorithms (\cite{such2017deep}). The diversity of their exploration, and its capacity to generate novel solutions from few human assumptions or information, have been relevant to model systems that show autonomous adaptation (\cite{holland1992genetic}, \cite{palmer1994artificial}, \cite{holland1999echoing}, \cite{lehman2020surprising}), coevolution (\cite{garcia2017investigating}) or learning (\cite{axelrod1987evolution}, \cite{lebaron1995experiments}, \cite{vie2020blotto}).  \par
These qualities are counterbalanced by severe constraints. First, genetic algorithms are often computationally costly to run. Their convergence speed often increases with higher population sizes, but this increase in the number of elements to evaluate may cause the computation cost in time and hardware to spike (\cite{deb2002fast}, \cite{deb2002nsga}). Several parameters, from population size to occurrence probabilities for the GA operators such as mutation and recombination rates, have decisive, non-trivial effects on the convergence properties and speed of the algorithm (\cite{grefenstette1986optimization}). Identifying proper sets of configurations across several parameters is a difficult task (\cite{de2020systematic}). Finally, the choice of the representation of solutions (\cite{ronald1997robust}), of the objective function (\cite{lehman2020surprising}), and of the method of creation of initial solutions (\cite{yakovlev2019optimization}) may cause biases on search, or be detrimental to computation efficiency, making these choices crucial and challenging steps.  \par
While these have been long standing issues in genetic algorithms from their introduction (\cite{holland1975adaptation}), recent research sheds new light on these challenges. The recent technological progress in computer hardware, with GPU (\cite{cheng2019accelerating}), Cloud (\cite{liu2017deadline}), quantum (\cite{malossini2008quantum}) and parallel (\cite{tang2017mapping}) computing, may allow to mitigate computational cost issues. Much progress has been done in parameter tuning and control, to determine respectively fixed and dynamic parameter configuration (\cite{huang2019survey}), allowing to more easily setup genetic algorithms. Examples include self-adaptation, in which parameters are themselves subject to evolution (\cite{dang2016self}, \cite{case2020self}), or covariance matrix adaptation (\cite{hansen2016cma}). New representation strategies have been proposed (\cite{vie2021sga}) and carry promising development, expanding the complexity of the objects GAs can evolve, and the difficulty of the tasks they may be able to perform. \\


By compiling these elements, qualities and challenges, with the addition of recent perspectives, this reviews attempts to inform newcomers and practitioners alike on genetic algorithms properties, and to highlight promising avenues for future research. Large potential discoveries may arise from "hybrid methods" associating GAs with more specialised optimisation algorithms, for example in deep neural networks (\cite{stanley2019designing}) or reinforcement learning (\cite{drugan2019reinforcement}). Such interdisciplinary endeavors may spark new discoveries in social sciences (\cite{chen2001evolving}, \cite{martinez2009heterogeneous}), open ended evolution (\cite{stanley2017open}, \cite{lehman2020surprising}), artificial life and AI (\cite{clune2019ai}, \cite{stanley2019designing}). \\


This review is structured as follows. We first introduce the genetic algorithm technique and its functioning in section \ref{genetic algorithms}. Their qualities for exploring large, complex search spaces, their optimisation qualities, and their close connection with emergence and novelty, are described in section \ref{qualities}. Key issues and challenges faced by GAs, in particular computation efficiency, parameter configuration and robustness, are presented in section \ref{problems}, that also presents recent perspectives on these issues. Final section \ref{conclusion} concludes on the qualities and challenges of genetic algorithms, and introduces promising perspectives related to renewed inspiration from biology and genetics, self-adaptation of GAs, open-ended evolution and AI.

\section{Genetic Algorithms}
\label{genetic algorithms}

Before exposing the qualities, limitations and perspective of GAs, we present in this section their core concepts, and their main steps. This section aims at presenting a reachable description of how genetic algorithms function, what the main concepts they rely on are. We first define key notions such as population, genetic representation, the relation between the genotype and the phenotype, and fitness, that acts as the driving force of evolution in GAs. We then describe the main algorithm steps, the main techniques used for each step, and the parameters involved. 

\subsection{Core concepts}

\subsubsection{Population and individuals} 

A genetic algorithm (GA) is a member of the family of \textit{evolutionary algorithms} (EAs), that are computational search methods inspired from natural selection. They simulate Darwinian evolution on individual entities, gathered in a \textit{population}. In the same ways giraffes evolved towards longer necks, or humans with bigger brains, evolution synthesised in a computer program strives to improve these entities. These individual entities can be anything: digital (i.e. artificial) organisms, bit strings, computer programs, financial strategies... As evolution affects the entities, subsequent versions of the population, also denoted \textit{generations}, evolve to maximise their fitness, i.e. ability to survive in an environment or to perform a given task, or to maximise a measure of performance. In the same way unicellular organisms evolved towards \textit{homo sapiens}, EAs turn unsophisticated bodies into complex, adaptive entities. As a computer program, all this evolution process starts from with a way to represent those entities: the \textit{genetic representation}.

\subsubsection{Genetic representation} 

GAs form a distinct sub-category of EAs by explicitly using a genetic \textit{representation} of those entities. That is, rather than mathematical functions or abstract objects, entities are formalised in the computer program in a way close to genetics, to a DNA sequence. The representation of the entities is a sequence of characters, that encode characteristics. How the entities characteristics or behaviors are formulated in the computer program is denoted a \textit{representation} (or \textit{encoding}). In GAs, these entities are represented with a \textit{genotype}, and a \textit{phenotype}, concepts from genetics. The \textit{representation} using a \textit{genotype} can be seen as encoding genetic information on their characteristics, in a simplified form of DNA. While several GAs do not make this distinction and confound genetic information and entity characteristics, it is inspiring and useful to briefly highlight these concepts. 

\paragraph{Genotype} The \textit{genotype} of an entity consists in its genetic information. It can be expressed in various ways. A popular representation technique in GAs is a bit-sequence with \textit{binary encoding}, i.e. composed of 0s and 1s. Similarly to the human DNA, genetic information can also be stored with an alphabet encoding, e.g. A,C,T,G as in DNA. The field of \textit{genetic programming} even considers genotypes constituted of sequences or trees or computer instructions, or mathematical operators. Notwithstanding this diversity of representations, genotypes are mostly encoded as a string, or sequence. These strings are often called \textit{chromosomes}, a sequence of elements denoted \textit{genes}, in analogy with the way genetic information is stored in living organisms. \cite{holland1992genetic} used the term \textit{schema} or \textit{building block} to describe some specific groups of genes, that are building blocks of the entity genotype. The location of genes in the chromosome is denoted as their \textit{locus}, and it is usually assumed that each gene has its own locus on the chromosome: in a sequence [1 2 3], each element is a unique gene present at its own separate locus. The different values each gene can take are denoted \textit{alleles}. For example, in the binary encoding, each gene can have either \textit{allele} 0 or 1. The binary chromosome becomes a bit string, i.e. a sequence of zeros and ones. In the above "DNA" alphabet encoding, each gene could take alleles A C T or G. Even though most living organisms have several chromosomes present in multiple copies, let us note that GAs often restrict to entities: that are \textit{haploid}: their genotype is characterized by only one chromosome; and \textit{monoploid}: these entities only have one copy of each chromosome. Figure \ref{genotypes} presents some examples of commonly used genotypes. \par

\begin{figure}[!htb]
\minipage{0.32\textwidth}
$$ \begin{pmatrix} 0 & 1 & 0 & 1 & 1 & \dots \end{pmatrix}$$
  \caption*{Binary genotype}
\endminipage\hfill
\minipage{0.32\textwidth}
$$ \begin{pmatrix} A & T & A & G & C & \dots \end{pmatrix}$$
  \caption*{Alphabet genotype}
\endminipage\hfill
\minipage{0.32\textwidth}%
\centering
\begin{tikzpicture}
\Vertex[x=1,y=1,size=.65,label=x,Math, color = white]{pt1}
\Vertex[x=2.4,y=1,size=.65,label=1,Math, color = white]{pt-1}
\Vertex[x=1.7,y=1.7,size=.65,label=-,Math, color = white]{-}
\Edge(pt1)(-)
\Edge(pt-1)(-)
\end{tikzpicture}
\caption*{Tree genotype (genetic programming)}
\endminipage
\caption{Instances of genotype representations. Each element of the chromosome (or sequence) is a gene at a specific locus. The number of genes is often defined as the \textit{length} of the chromosome.}
\label{genotypes}
\end{figure}

\paragraph{Phenotype} The \textit{phenotype} corresponds to to the entity characteristics: shape, size, color, abilities... that are encoded in the genetic information (genotype). The phenotype maps this information to specific behaviors, the \textit{traits} of the entities. For instance, genes OCA2 and HERC2 located on chromosome 15 in humans are considered as the main genes responsible for the eye color trait. In this example, the genetic information contained in genes OCA2 and HERC2 is the genotype. The resulting eye color, and the way the alleles of OCA2 and HERC2 correspond to a specific color outcome, belong to the phenotype. \par
\textit{Genotype} and \textit{phenotype} differ: while the phenotype can be observed, the genotype cannot. But one crucial point to add here is the role of the \textit{environment}. It is usually considered that the genotype and the entity environment together determine its phenotype, rather than genetic information alone. This aspect is often not present in past and current GAs. They usually consider the correspondence between genotype and phenotype to be fixed regardless of environmental contexts. We will return in more detail to this consideration. We have now introduced the way entities are represented in our evolution program. As in nature, the driving force of their evolution is the fitness. 

\subsubsection{Fitness} 
In natural systems, evolution favors entities that are more capable of reproducing, due to some advantages they have: they could be stronger, they could be more attractive, they could be more adapted to their environment. In GAs, fitness measures the performance of the entities in the population, in adapting to a given environment or performing a given task. In natural evolution, fitness criteria are implicit. Entities struggle to reproduce and to survive, by all means available, which can be described with a simulation. In most GAs however, the fitness criterion is made explicit by the introduction of a \textit{fitness function}, that allows to rigorously associate phenotypes to fitness. This is closely related to similar concepts of rewards in reinforcement learning, payoffs/loss function in general machine learning, or value function in optimisation. These fitness functions may be simple, such as counting the number of ones in a bit string (known as the Counting Ones problem). Other may be much more complicated, such as optimising the returns of a complex trading strategy, or solving a multi-objective optimisation problem. These fitness functions define implicitly or explicitly a \textit{fitness landscape} (\cite{wright1931evolution}). \textit{Fitness landscapes} are a physical representation of the relationship between phenotypes and fitness. Some phenotypes may enjoy a very high fitness, corresponding to peaks, or mountains in the fitness landscape. Some behaviors may have low fitness, and would appear in the landscape as valleys or caves. One important distinction between GAs and their machine learning cousins is that by evolving the genotypes of the entities in the population, the GA maintains a population of phenotypes exploring the fitness landscape guided by evolution-inspired operators, rather than a unique \textit{gradient-descending} individual (that moves in the direction of the largest improvement, as measured by the derivatives of the fitness function). The evolution of the population strives to identify phenotypes with good fitness: escaping valleys and climbing mountains. By so doing, the GA explores the \textit{search space}: the space of possible solutions or entities. How large, \textit{smooth} or \textit{rugged} this fitness landscape is, and how difficult this exploration process can be, will depend on the specific problem being considered, and the genetic representation being used. By \textit{smooth}, we will speak of a fitness landscape that is single-peaked, or \textit{uni-modal}. Otherwise, it may have multiple mountains (\textit{multi-modal}), or "hide" a mountain in a low valley, or present some irregular structures: this characterises a \textit{rugged} fitness landscape. The next Figure \ref{landscape-example} shows graphical examples of such simple and rugged fitness landscapes to make these concepts clear.

\begin{figure}[h]
    \centering
    \includegraphics[scale = 0.25]{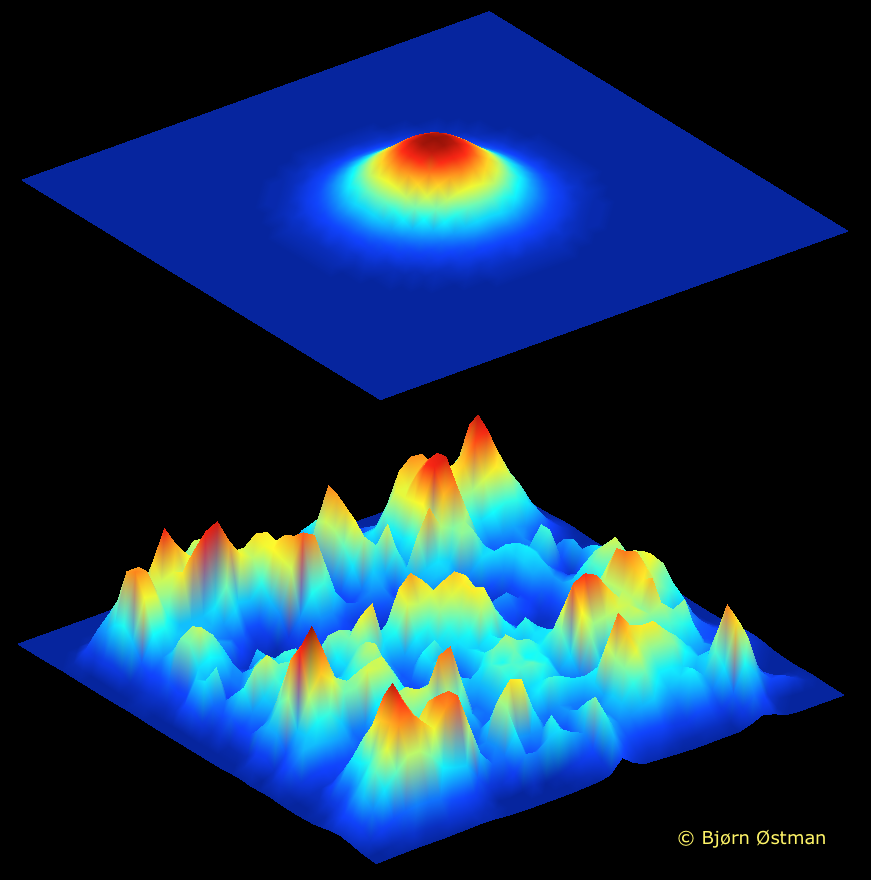}
    \caption{Examples of a simple, smooth (top) and rugged (bottom) fitness landscape (\cite{ostman2013blog,ostman2014predicting}). Peaks and mountains that correspond to high fitness are colored in red, valleys of low fitness in blue. The fitness landscape \textit{ruggedness} is associated with the irregularity of the landscape, its number of different peaks. Rugged fitness landscapes are thus more difficult to explore.}
    \label{landscape-example}
\end{figure}

One could see the GA as a population of hikers that try to find the highest mountain of the country. At first, they are randomly placed in the fitness landscape, and only see the very small territory in which they are placed. Each has its own strategy for hiking: some will go north, others will go south, some decide to move randomly, others want to run... These hikers communicate with each other, combine their information on their positions, and move across the landscape trying to find a higher mountain to climb. Even though each evolves in its own interest, they develop collective intelligence by using the other hikers' information and strategies to improve their own. If an agent finds a huge mountain walking north, some others agents are more susceptible to include a north direction to their movement. By so doing, this decentralised population of hikers achieves a strong ability to explore large, rugged landscapes, and to quickly identify good regions, that is, good points in this landscape. This exploration is intuitively harder when there are multiple mountains of different size, and valleys in between. But the collective nature of the search make GA capable of reaching the highest mountain nevertheless.

\subsection{Algorithm steps}

Genetic algorithms essentially proceed with five main steps, that are present in all GAs, though with various implementations. First, an initial population of entities is generated: the first generation. Then, the following steps are repeated for a \textit{number of iterations}. The entities in the current generation are evaluated in terms of fitness. Some parents are selected in this generation to reproduce. By reproducing, they create offspring that share some of their genetic information. The resulting children genotypes are affected by mutations that increase population genetic diversity. Iterating this procedure until we have a complete new generation constitutes a \textit{step }of the GA. Once that new generation is created, we iterate the program steps until this evolution has reached some objective, or until a maximum number of iterations has been reached. 

\subsubsection{Creation of the initial population}

The GA usually starts from a random population of chromosomes entities, whose number is the \textit{population size}. The objective of this initial sampling is to spread uniformly the initial solutions in the search space. Covering the space of possible entities as uniformly as possible aims at being more likely to quickly find promising regions of the fitness landscape (\cite{mirjalili2019genetic}), and to minimise bias in search. For example, under the binary encoding, we will typically generate random sequences of 0s and 1s as our initial population. 

\begin{figure}[H]
$$\underbrace{\begin{pmatrix} 0 & 1 & 0 \\ 1 & 0 & 1 \\ 0 & 0 & 0\end{pmatrix}}_{\text{population size of 3}}$$
\caption{An example of random initial sampling for a population of 3 individuals, in the space $\{0,1\}^3$.}
\end{figure}

This requirement of unbiasedness is key to utilisation of GAs as optimisers. For the evolved solution to be robust and global, maximising the diversity of candidates in the first generations is a priority. Maintaining this diversity of entities points in the search space allows to prevent convergence to local optima. However, other uses of GAs may relax this condition. Studying how specific bias in initial populations affect their evolution may be a valid research question on its own, and an interesting experiment to understand the functioning of GAs. 

\textit{Population size} an important parameter. A higher population size allows to maintain more diversity, and improves natural selection in the population. It however makes the search slower, and computationally more costly. In most GAs, the population size is fixed, and restricted to a few tenths or hundreds of individuals. Yet, in living systems, population sizes are very large, and their sizes are the product of complex ecological interactions.

\subsubsection{Fitness evaluation}

Evaluating the fitness of all entities in the current generation consists in applying our fitness criterion to them. As we described, this could take the form of a simulation, or of a function, depending on how fitness was defined. Regardless of this choice, the fitness evaluation procedure assigns to each entity a fitness score. This procedure allows to discriminate between entities with low and highly fit phenotypes (as an indication of low or highly fit genotypes), which becomes essential in the selection of parents to build the next generation of entities. As genetic information will be inherited by children entities, we aim at selecting good, diverse genetic material in the parents organisms. \par

Four measures of fitness are usually distinguished. \cite{koza1992genetic} notably sets a clear difference between the \textit{raw fitness} and other adjusted measures. \textit{Raw fitness} $r$ is the exact measure of the fitness function, that is natural to the problem terminology, i.e. error rate, or number of ones... \textit{Standardised fitness} $s$ can be used to restate a fitness maximisation problem in a minimisation problem, by setting $s = r_{\text{max}} - r$ where $r_{\text{max}}$ is the largest possible raw fitness value. \textit{Adjusted fitness} $a = \frac{1}{1+s}$ creates a measure in $(0,1)$ of fitness between individuals that allows to discriminate entities with only small differences in fitness. Finally, \textit{normalised fitness} $n=\frac{a}{\sum a}$ allows to measure the relative fitness of an entity with respect to the fitness of the whole population. These different measures allow different selection methods that we present below.

\begin{figure}[H]
$$\begin{pmatrix} 0 & 1 & 0 \\ 1 & 0 & 1 \\ 0 & 0 & 0\end{pmatrix} \text{ gives } \begin{pmatrix} 1 \\ 2 \\ 0 \end{pmatrix}$$.
\caption{An example of raw fitness evaluation. The fitness criterion measures how close the individuals are to $(1 \ 1 \ 1)$.}
\end{figure}

\subsubsection{Selection of parents}

Once we have identified the fitness of each entity, we now select some as parents. Two parents will mate, exchange genetic information, and generate two offspring that share some genetic information from their parents. The process of selection of parents continues until we have enough offspring to constitute the next generation. A competitive selection of parents allows transmission of fit phenotypes and genotypes in the next population. A large variety of selection methods do exist to reproduce this reproductive advantage phenomenon. \\

Several selection methods are based on one of the four measures of fitness. The \textit{fitness proportionate "roulette-wheel"} selection method is very frequent (\cite{mirjalili2019genetic}). Each entity has a probability to be selected as parent, and this probability depends on the ratio of its fitness to the cumulative fitness in the population, that is, the \textit{normalised fitness}. The more an entity stands out in the population for its fitness, the more likely it is to be selected as parent. Draws are made without replacement, so a highly fit individual could reproduce several times, and transmit its genotype more. \textit{Elitist selection} often complements other selection methods, as an additional layer. It consists in saving without modification the current best entity in the next generation (\cite{ahn2003elitism}), in order to save the current best individual during search. Others selection methods are based on rank, and select a given number of entities ranked by their fitness values. \par 
\textit{Rank selection} (\cite{razali2011genetic}) proceeds likewise. With a parameter that specifies the number of individuals to select, rank selection selects as parents this number of best entities indexed usually by \textit{adjusted} or \textit{standardised fitness}. Different methods attempt to measure Pareto dominance between entities, i.e. establish ordinal comparisons. This is the essence of \textit{tournament selection} (\cite{miller1995genetic}). In this approach to parents' selection, a subset of the population is selected, each each entity in this subset is compared with the others members of the subset to determine the best individuals. The winners of these tournaments become parents. \\

The success of GAs as explorers and engines of successful evolution relies on a delicate balance between selection pressure and diversity. We desire to select the best entities to improve the next generation, without looking only at short term rewards and narrowing our search. Both can be adjusted by the selection of the selection method, or some related parameters such as the size of the tournament in tournament selection. A tournament against all the other entities of the population puts a strong selection pressure, and unfit entities have no chance of passing through the selection step. If the tournament is only organised against a few other entities, a lucky draw may allow a weaker entity to be selected as parent. But to guarantee that this process does not fall short, and only admits short term rewards or local optima, some diversity must be kept in the population. What is the optimal selection pressure, and diversity level, depend on the phenomenon or problem being considered. \par
Alternative selection methods such as truncation selection, sigma scaling, local selection proportional selection, are reviewed by \cite{mirjalili2019genetic}, and follow this attempt to finely tune selection pressure to the specific problem being considered. In their Niched Pareto GA, \cite{horn1994niched} introduced Pareto tournaments, selecting two individuals and $n$ others in a comparison set on which to evaluate dominance. Concerns on maintaining balance and exploration have encouraged the development of selection methods rewarding diversity, such as crowding sharing rules (\cite{horn1994niched}), the "1/4+" selection method, explicit fitness-sharing in genetic clusters (\cite{goldberg1987genetic}), novelty search (\cite{lehman2011abandoning}), or quality-diversity algorithms (\cite{pugh2016quality}). The latter uses rank rather than fitness score to operate selection, and operates a random selection in the top 20\% of the population. This is used to prevent stronger individuals from quickly dominating the population, driving the genetic diversity down too early (\cite{packard1988adaptation}). 

\subsubsection{Reproduction and recombination of parents' genotypes}

After selecting promising parents based on fitness or rank, recombination step combines their genetic material to create two new entities in the population. This is an essential step of the GA, in which self-replicating entities transmit their genotype to the next generation, ideally saving good characteristics, and discarding bad characteristics. These offspring share a combination of the genetic information of their two parents, which allows them to explore new regions of the search space -that is, new phenotypes-, taking advantage of previous evaluations. With the idea that two performant parents have been selected, recombination allows to mix their genotypes together to evolve a potentially more performant child. \par

\begin{figure}[H]
\centering
\begin{subfigure}{.5\textwidth}
  \centering
  \includegraphics[width=\linewidth]{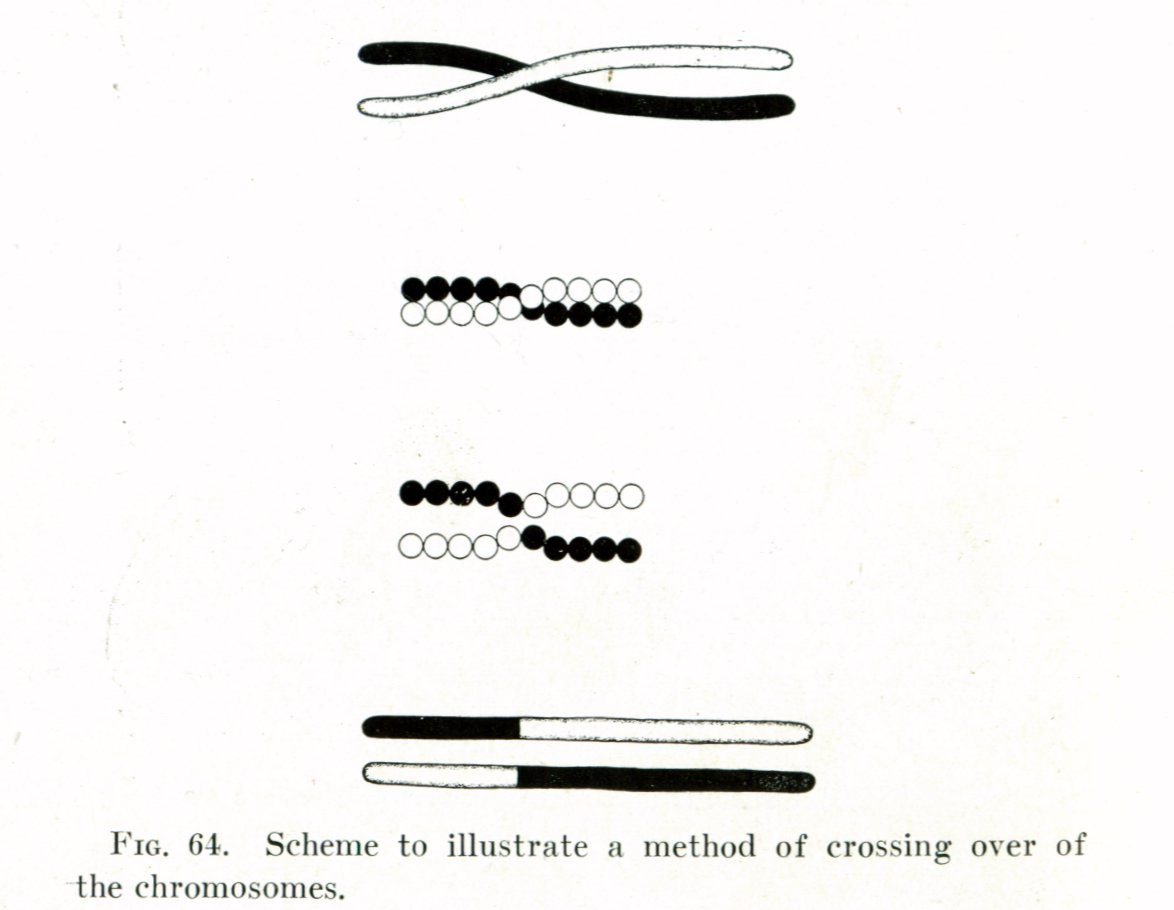}
  \caption{Single-point crossover}
  \label{single-crossover}
\end{subfigure}%
\begin{subfigure}{.5\textwidth}
  \centering
  \includegraphics[width=\linewidth]{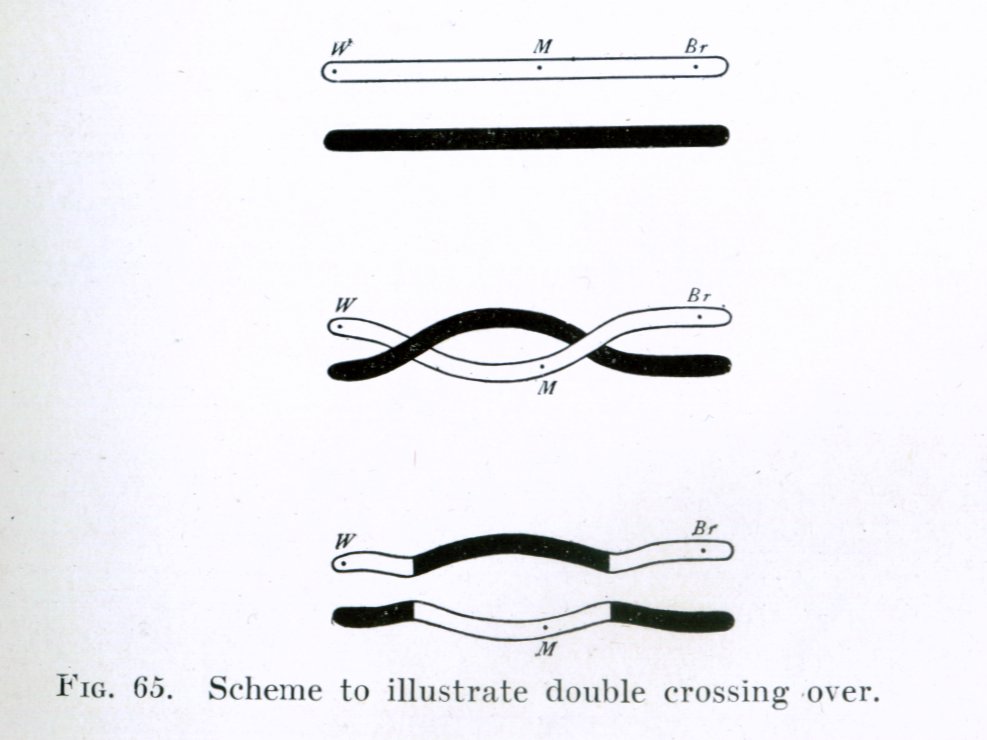}
  \caption{Double-point crossover}
  \label{double-crossover}
\end{subfigure}
\caption{Two crossover mechanisms. Illustrations from Thomas Hunt Morgan (1916).}
\label{crossover}
\end{figure}

Most of the literature in GAs is inspired from the \textit{single-point}, or \textit{double-point} crossover, even though several variants and alternatives do exist (\cite{mirjalili2019genetic}). Considering again \textit{monoploid} entities, with parent Black and White, their chromosome cross at a given point: the \textit{crossover point}. This point is unique in single-point crossover, and could be at any locus. Its \textit{locus} is typically chosen at random, according to a uniform distribution. Each locus is as likely as the others to be the crossover point. This is denoted by the terminology \textit{single-point uniform crossover}, illustrated in Figure \ref{single-crossover}. A frequent alternative is \textit{double-point uniform crossover}, that simply adds a second crossover point to make the recombination finer, as illustrated in Figure \ref{double-crossover}. Figure \ref{crossover-binary} provides an instance of single-point uniform crossover in a binary encoding environment. These methods are very natural in GAs, as they represent entities with a sequence of genes. 

\begin{figure}[H]
$$\left(
\begin{array}{c|cc}
\textcolor{violet}{0} & \textcolor{violet}{1} & \textcolor{violet}{0}\\
1 & 0 & 1\\
\end{array}
\right) \longrightarrow \left(
\begin{array}{c|cc}
\textcolor{violet}{0} & 0 & 1\\
1 & \textcolor{violet}{1} & \textcolor{violet}{0}\\
\end{array}
\right) $$
\caption{An instance of a random uniform single-point crossover illustrated by the bar. Genes are exchanged after the crossover point, returning the genotype of their 2 offspring. }
\label{crossover-binary}
\end{figure}

This crossover mechanism is the most frequent implementation of the recombination step. It borrows from biology in genetic recombination, and corresponds to the process called synapsis during \textit{meiosis}, the cell division of germ (reproductive) cells. The swapping of genetic information creates genetic variation and diversity. In GAs, this diversity is controlled by a probability that the crossover occurs. This is denoted the \textit{crossover probability} or \textit{crossover rate}. If this rate is equal to 0.9 for instance, this is the probability that the two offspring are generated by the above process. In 10\% of the cases, children genotypes are the exact copies of the parents, without recombination. This allows to control the stability or diversity of the population during search, to make the algorithm faster or more robust during search, depending on the constraints of the problem.

\subsubsection{Mutation of children' genotypes} An additional way to maintain this diversity is the \textit{mutation} step. Maintaining diversity over generations is essential to GA performance (\cite{horn1994niched}). GAs allow some random perturbations of the genotypes, leading to beneficial, neutral or detrimental phenotype variations. They are considered as an important source of genetic diversity, and can transmit to the next offspring of a mutated entity. They are then part of the exploration and evolution strategy. An individual with a beneficial mutation enjoy a fitness advantage, making it more likely to reproduce and to transmit this beneficial trait to the next generations. Conversely, an entity receiving a detrimental mutation tends to be discarded, and the trait may disappear in the next generations. Mutations also provide an insurance against the development of uniform populations incapable of further evolution (\cite{holland1992genetic}).\par

Mutations correspond to perturbations in the genotype sequence. One common example in GAs is the \textit{flipping mutation}, or \textit{point mutation}. In the human DNA, an A can turn into a C, changing the encoded basis, protein activation, with possibly consequences on the phenotype. In a binary string, a zero might change into a one or a one to a zero, along the genotype sequence. Figure \ref{mutation-binary} presents an example of such mutation. Mutations occur independently for each gene regardless of position or previous mutations. Any locus has an independent and identical probability of mutation, usually assumed to happen with a uniform probability: the \textit{mutation probability}. All loci of all entities in the population can be potentially affected, as all mutation draws are usually independent between loci, and between individuals. The determination of the mutation rate is important, as a null mutation probability harms search by reducing diversity, but a too large mutation rate may disrupt good genotypes being evolved. In nature, these exact mutation probabilities are difficult to estimate, but range in the order of $10^{-8}$.

\begin{figure}[H]
$$
\left(
\begin{array}{ccc}
0 & 0 & 1\\
1 & 1 & \textcolor{orange}{0}\\
\end{array}
\right) \longrightarrow \left(
\begin{array}{ccc}
0 & 0 & 1\\
1 & 1 & \textcolor{orange}{1}\\
\end{array}
\right)  $$
\caption{An example of a uniform flip mutation on the third gene of the second child.}
\label{mutation-binary}
\end{figure}

A wide diversity of implementations of mutation mechanisms exist in GAs (\cite{mirjalili2019genetic}), but not as much as in nature. These mutations operators vary with the problem being tackled. To handle dynamic fitness landscapes in which the optima change over time, \cite{grefenstette1992genetic} introduced a \textit{partial hyper-mutation} operator, which essentially replaces a given percentage of the population with random entries. This allows the GA to maintain a continuous level of exploration of the search space. Natural systems are rich with other instances of mutations. \cite{lindgren1992evolutionary} introduced a \textit{duplication} mutation that extends the genotype length. Genes may be \textit{shuffled} or \textit{inverted}. Some sections may be subject to \textit{deletion}. Some genes may be subject to \textit{translocation}, and be interchanged with in another locus by \textit{substitution}. 

\subsubsection{Iteration and evolution} From the creation of the initial population, the repetition of these four steps: evaluation, selection, recombination and mutation, allow the population of entities to explore the search space, identifying promising genotype sequences and phenotypes with respect to the fitness criterion. Recombination and mutation of the best entities allow an "intelligent" search, that uses previous evaluations to decide what points of the search space to exploit next, while maintaining diversity and continuous exploration. The question of the \textit{terminal condition} of the GAs is often discarded. Most of the GAs are run for a predefined \textit{number of iterations} (repetitions), after which the outcome and the population trajectory are observed. When we are solving a problem with a known answer, we can terminate the iteration of the above steps once the known solution has been identified. But as we would typically use GAs to learn something we do not already know, it is not trivial to guess how many iterations will be necessary, or to anticipate how different number of iterations would yield different outcomes.\par
The other parameters involved in these steps allow to tune selection pressure, stability of the populations, diversity, in order to generate a robust process that identifies and converges to the global optimum in the fitness landscape. These operators and steps may seem basic. They are indeed simplifications of the real biological processes at play. However, they have nurtured significant achievements from such simple approximations that we will now describe. Now that we presented an overview of the definition, concepts and steps of the genetic algorithm, from initial sampling to terminal conditions, let us now move to enumerating the merits and drawbacks of this approach. 

\section{Evolving good solutions in complex environments}
\label{qualities}

First, GAs achieve an optimal balance between exploration and exploitation that places them as powerful explorers of complex landscapes, large and rugged. Second, this exploration quality places GAs as a powerful, robust optimisation tool for solving difficult problems: either admitting a large number of possible solutions, facing difficult constraints, or being too complex to be solved in reasonable time with traditional methods. Finally, the diversity of entities evolved by GAs place them as adequate techniques to model emergence, simulate the behavior of evolutionary systems, and generate novelty.

\subsection{Genetic Algorithms as powerful complex landscapes explorers}

\subsubsection{Exploring large search spaces}
In a stark contrast with exhaustive, enumeration or random search-based methods, GAs are suitable to solving \textit{sparse problems}. That is, problems for which the number of good solutions is very small with respect to the search space size (\cite{whitley1994genetic}, \cite{maulik2000genetic}). GAs evolve a multitude of genotype sequences over many regions simultaneously. By their ability of combining sequences containing good partial solutions, GAs quickly focus their attention on the most promising locations of the search space (\cite{holland1992genetic}). Recall our analogy on the population of hikers acting collaboratively. The GA places a population of points in the fitness landscape, and simulates a collective, decentralised search in this landscape. Having collaboration between hikers contributes to improve efficiency, mutations bring diversity, and the population features allow to simultaneously explore different high fitness regions of the fitness landscape, that correspond to good solution points in the search space. \par

GAs in such very large, high-dimensional spaces, have reached significant performance in pattern recognition and clustering in artificial and empirical data sets, in various dimensions and number of cluster ranges (\cite{maulik2000genetic}). GA-clustering algorithms notably outperformed than the traditional K-means algorithm. Similar performance was achieved in computational protein design (\cite{street1999computational}) molecular geometry optimization by \cite{deaven1995molecular}. Determining the lowest energy configurations of a collection of atoms is \textit{NP-hard} (a class of problems that notably require a long -polynomial- time for solving) and covers a very large space of possible solutions, as the number of candidates scales exponentially with the number of atoms. Nevertheless, the GA quickly found the best structure, in spite of strong directional bonds between different structure. By so doing, the GA outperformed simulated annealing (SA). In large flow-shop sequencing problems, GAs reached near-optimal solutions more quickly than SA (\cite{reeves1995genetic}), or than local discriminant analysis (\cite{varetto1998genetic}), showing its relevance for large and difficult combinatorial problems. \cite{packard1990genetic} evolved prediction models in financial markets. Searching through the -huge- space of sets of conditions and predictions, the algorithm identified regions of predictability in stock market data, and obtained robust results in out of sample forecasting. \par

These large spaces of possible solutions appear frequently in strategic games. \cite{axelrod1987evolution}'s GA application to the Prisoner's dilemma efficiently scanned over $2^{64}$ (16 quadrillion)) strategies, exploited some weaknesses in sample strategies to perform best than the optimal tit for tat rule (cooperate unless the opponent has defected, then defect), and developed tit for that on its own. Recently, \cite{vie2020blotto} used a GA to identify optimal strategies in completely open environments that admitted countably many alternatives, and \cite{mirjalili2020genetic} reconstructed images based on random samples.  
Whether they explore the set of prediction rules (\cite{waheeb2019new}, \cite{han2019using}), of computer programs (\cite{devarriya2020unbalanced}), of deep neural network architectures (\cite{chung2020genetic}), strategies in games (\cite{axelrod1987evolution}, \cite{vie2020blotto}), the space of different combinations of portfolios (\cite{lwin2014learning}), economic behaviors (\cite{chatterjee2018efficient}), the search spaces involved are often very large. Most of them also exhibit \textit{rugged} structure as in Figure \ref{landscape-example}: the path to the global optimum may not be smooth, several (local) optima may exist. GA are nevertheless performing well in such complex fitness landscapes.

\subsubsection{Exploring rugged search spaces}

A traditional technique to explore and optimize in landscapes is \textit{hill-climbing}, or variants of \textit{gradient descent}: start at some point, and follow the path to the greater improvement to the quality of the solution. However, as the landscape becomes more rugged, complex, irregular, with many high points, tunnels, bridges or even some more convoluted topological features, finding the right hill and which way to go becomes increasingly hard (\cite{holland1992genetic}), or may not converge to the global optimum, getting stuck in local optima. GAs do not require the optimisation landscape to be differentiable (a necessary condition for using gradient descent), uni-modal or single-peaked (\cite{maulik2000genetic}) or to not exhibit particular topologies. Maintaining a population of solutions rather than a single solution, GAs are less vulnerable to premature convergence to local optima, as long as a sufficient level of population diversity is maintained (\cite{maulik2000genetic}, \cite{Wirsansky2020hands}). Some examples illustrated below (from \cite{Wirsansky2020hands}) are the optimization of the Eggholder function, known for its large number of local extrema, and for which we can analytically derive the global minimum, and of the Himmelblau's function that admits four global minima. In these cases, maintaining diversity in the population allows to find the global optimum (or optima) without being stuck in local optima, illustrating nice properties of GA search, but also configuration requirements to guarantee these search qualities.

\begin{figure}[H]
\minipage{0.49\textwidth}
    \includegraphics[width = \textwidth]{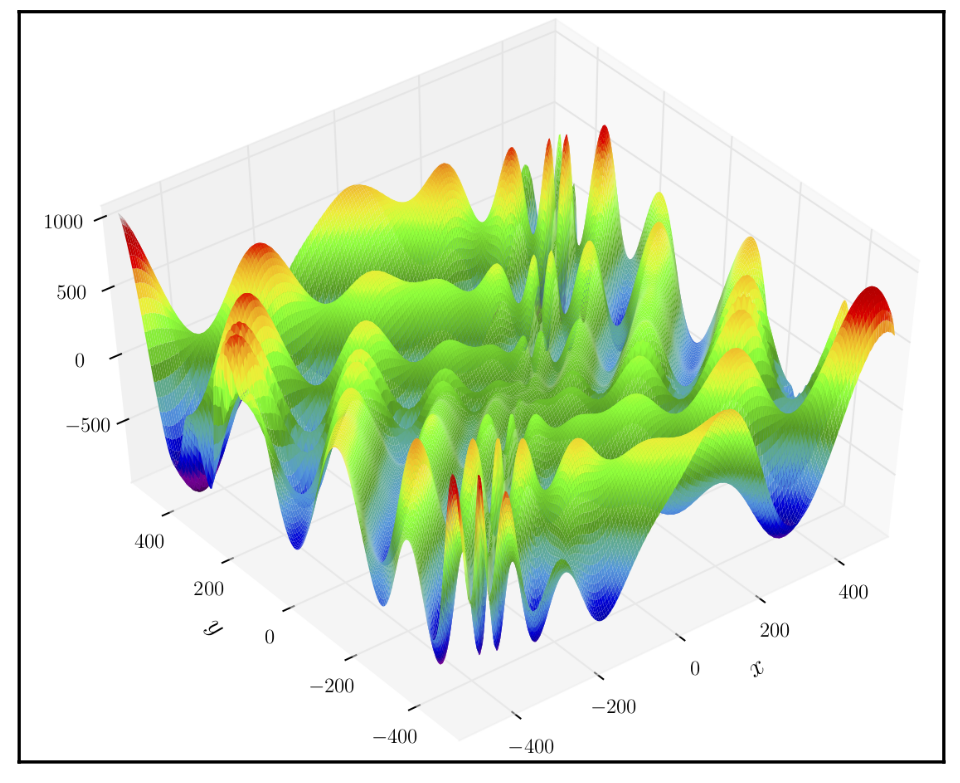}
    \caption{The Eggholder function}
    \label{eggholder}
\endminipage\hfill
\minipage{0.49\textwidth}
    \includegraphics[width = 0.965\textwidth]{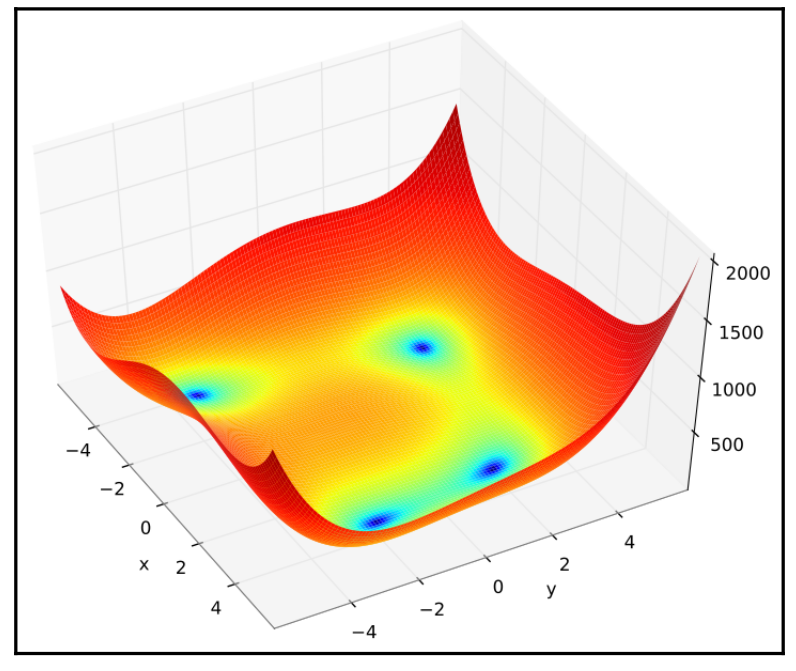}
    \caption{The Himmelblau function}
    \label{himmelblau}
\endminipage
\label{eggholder_himmelblau}
\end{figure}

When dealing with multi-modal fitness landscapes in which several global extrema coexist, GAs have often been criticised for converging to an arbitrary instance of global extrema due to \textit{genetic drift}, and \textit{sampling bias} during search, rather than maintaining a stable distribution of the population around the different optima. \textit{Genetic drift} describes a situation in which the population converges to one of equally good traits arbitrarily, because of noise or chance. \textit{GA sampling} describes the process of generating an initial distribution. This issue is solved by introducing randomness and different GA sampling, to obtain asymptotic unbiasedness and the adequate distribution of solutions as the number of GA runs increases: see notably such an application for the optimization of the Himmelblau's function (\cite{Wirsansky2020hands}). This means that as we run different instances of GAs with different initial populations, we can converge to the true result, being confident that this result is not the outcome of a specific starting condition. Alternatively, the introduction of niching sharing rules that penalise similar solutions to avoid clustering of solutions in only one of the extrema is an adequate means to prevents genetic drift. This essentially is analog to increasing diversity during search. For a single-peaked problem, the resolution is still possible but may take longer, but this makes the program capable of handling multi-peaked problems. These rules allow convergence to the four minima of the Himmelblau's function in a single run (\cite{Wirsansky2020hands}). So far, we have presented two important areas in which GA thrive: problems with large search spaces, and for which the fitness landscape associated with the problem is rugged. This naturally places GAs as robust solvers of complex problems.

\subsection{Solving complex problems}

In the continuity of their exploration and optimisation qualities, GAs provide a robust approach for complex problems, because of their flexibility and of their very low requirements for information on the problem.. Some major developments of GAs as problem solvers have taken place in the evolution of deep neural networks, multi-objective optimization, and more generally problems that are not fully understood by their decision maker.

\subsubsection{A robust method for optimization}

GAs are recognized as a "weak" or "robust" method (\cite{whitley1994genetic}) for their applicability to a high diversity of problems, and their ability to identify good candidates with very few assumptions made on the solutions of these problems. In problem solving, the entities evolved by the GA are solutions to this problem: some values to maximise a function, some classifier system, some sequence for scheduling, some list of parameters that encodes a function or an algorithm. When we describe the GA as a \textit{weak} or \textit{robust} method, we mean that the GA is relevant to tackle all these problems: it is a generalist method that is widely applicable. 
\par 
In matters of optimisation, previous mathematical research has established guarantees of convergence. Viewing a GA as a finite-state Markov chain, \cite{bhandari1996genetic} showed that that the canonical GA delivers the optimal string as the number of iterations goes to infinity, with a nonzero mutation probability. In a related fashion, \cite{del2001asymptotic} modelled GAs as a system of interaction particles with competitive selection and mutation, showing that the population asymptotically converges to a constant distribution corresponding to the global optimum of the fitness function. In well behaved landscapes, genetic algorithms results matched gradient type methods outcomes, in particular for nonlinear constrained optimization (\cite{homaifar1994constrained}). This demonstrates that matters of speed put aside, GAs evolve the same solutions as more traditional methods, while being operational at a broader scope. \par

This convergence result was initiated by Holland's schema theorem. \cite{holland1992genetic} states that good solutions -genotypes- to a given problem tend to be composed of good \textit{schemas}, i.e. building blocks: individual or groups of genes alleles. As the GA is evaluating some number $n$ of genotypes, it is by so doing evaluating the fitness of a much larger number of schemas. By evaluating the fitness of the string $(111)$, we are evaluating several axis of the search space: we are measuring the value of the building block 1 in each element. Both $(111)$ and $(110)$ are useful measures to evaluate if the schema $(11*)$ is a good strategy (in which $*$ can be either 0 or 1). This is the \textit{implicit parallelism} of the GA. Evaluating one genotype corresponds to evaluating all the building blocks present in the genotype, and by so doing, identifying the best schemas. Holland argues that what the GA optimizes generation after generation, are not genotypes, but those schemas. The Schema Theorem states that short, low order schemas (i.e. small sequences such as $(1**)$) whose fitness is higher than average will occupy an exponentially increasing share of the population over time. In this perspective, crossover operations aim at combining good schemas together to create even better higher order schemas. By so doing, the GA is capable of extracting meaning from noisy or imprecise information, or to detect trends that are too complex to be noticed by humans or standard techniques (\cite{metawa2017genetic}). \par

GAs can develop good solutions to problems we do not fully understand. For very sparse problems -recall: problems that have a small number of good solutions with respect to the number of possible solutions-, the efficiency of the GA algorithm was improved when the class, or "shape" of desired solution was assumed (\cite{deaven1995molecular}). GAs are however also relevant in problems for which we typically have little information on what optimal solutions would look like, and in which exploration of diverse candidates is desirable. In analysing insolvency risk with a GA, for which many different factors may be relevant, \cite{varetto1998genetic} showed that the GA obtained results faster than local discriminant with more limited contributions from the financial analyst. More recently, this low-information requirement has been useful for heart disease diagnosis (\cite{reddy2020hybrid}), design of neural network architectures (\cite{suganuma2017genetic}), production scheduling (\cite{nguyen2017genetic}), classification tasks such as detection of video change (\cite{bianco2017combination}), or determination of new functions to detect breast cancer (\cite{devarriya2020unbalanced}). Those are typical examples of problems for which we do not know precisely how the optimal solution looks like for designing a neural network for a specific classification problem, or what factors should be taken into account to best identify heart disease. GAs offer a competitive alternative to identify good solutions to such problems that we do not fully understand. We specifically describe some of the domains in which GAs have flourished. \par

\subsubsection{Evolving neural networks by exploring learning architectures}

GAs have been successfully involved in hybrid methods that combine GAs with other approaches (\cite{whitley1994genetic}), such as k nearest neighbor classification in dimensionality reduction (\cite{raymer2000dimensionality}). Quite early on, GAs have been successfully applied to feature subset selection (\cite{yang1998feature}) and dimensionality reduction (\cite{raymer2000dimensionality}). This preceded further combinations with neural networks, with very recent successes. The 1990s saw the first works on evolving network architecture, weights, or even the learning rules used by neural networks (\cite{kitano1990designing}, \cite{kitano1994neurogenetic}). Years later, GA-assisted neural networks have shown superior performance compared to fixed geometry neural nets and traditional nonlinear time series techniques in modelling daily foreign exchange rates (\cite{waheeb2019new}), daily bitcoin prices (\cite{han2019using}). \cite{chung2020genetic} used an hybrid GA with a convolutional neural network (CNN) that outperformed other models. They used GAs to identify optimal CNN architecture, an important determinant of the performance of these programs. A large number of recent works have applied GA methods in combination with machine learning techniques: traditional neural networks also named Evolutionary artificial neural networks (bank performance prediction: \cite{ravi2008soft}, cotton yarn quality \cite{amin2013novel}, time series forecasting: \cite{donate2013time}, machine productivity: \cite{azadeh2015flexible}, modelling of cogeneration processes: \cite{braun2016neuro}, air-blast prediction: \cite{armaghani2018airblast}, controller design: \cite{abd2018load}), support vector machines (parameter selection \cite{zhao2012parameter}, intrusion detection: \cite{ahmad2014enhancing} and \cite{raman2017efficient}, hospitalisation expense modelling: \cite{tao2019ga}) and case-based reasoning (corporate bond rating: \cite{shin1999case}, bankruptcy prediction: \cite{ahn2009bankruptcy}). Some recent breakthrough in reinforcement learning research have included some key ingredients of evolutionary algorithms. The pioneering AlphaStar algorithm of Deepmind (\cite{vinyals2019alphastar}) used the AlphaStar league, that can be described as a population of versions of AlphaStar fighting against other versions to improve and find new niche strategies. These achievements can thus be related to key concepts of evolutionary computation (\cite{arulkumaran2019alphastar}): Larmarckian evolution (driven by inheritance of acquired characteristics), competitive coevolution quality-diversity, that GAs naturally incorporate. Recently, the AutoML-zero algorithm evolved multilayer machine learning programs from scratch, only driven by evolution operators (\cite{real2020automl}). Computer simulations of evolution as done in these programs, and also by GAs that naturally include these mechanisms, hence constitute an already successful, but yet still further promising, area of research.

\subsubsection{Multi-objective optimisation}

\textit{Multi-objective optimization} (or multi-criteria) is a branch of optimisation that takes care of decision problems with several objective functions, and typically several constraints or decision variables. For instance, the problems analysed by \cite{vie2019long}, \cite{colapinto2020toward} and \cite{liuzzi2020sustainability} are multi-objective problems, because they involved a maximisation of workers allocations in various economic sectors in order to simultaneously maximise economic growth, employment, minimising electric consumption and greenhouse gas emissions. This is an instance of a problem that can quickly become very complex to analyse, as small variations in the solution can have mixed effects on the different objectives. While the above works have notably constrained the problem to linear constraints to be solvable, GA can handle more realistic settings and cope with uncertainty. \par

For this reason, this domain of multi-objective optimisation has been an important area of applications of GA as optimizers (\cite{horn1994niched}). Notably, \cite{lwin2014learning} obtained an algorithm capable of designing high quality portfolios in spite of numerous constraints, handling more than a thousand different assets, and outperforming state of the art programs. It is not surprising other applications moved to multi-criteria decision making with environmental factors: optimal energy storage units placement (\cite{ghofrani2013framework}), wind farm development (\cite{dhunny2020economic}), dynamic greenhouse environment control (\cite{jin2020engineering}), car engine design (\cite{tayarani2014meta}). Multi modal multi-objective optimisation has been an area of intense research in evolutionary computation more generally (\cite{8684299}). Designing supply chain networks facing multiple constraints, such as multi stages, multi products, multi plants with maximum capacities, has also been an area of success of GA applications (\cite{altiparmak2009steady}). In this case, their steady state GA was shown capable of handling dynamic environment, and stochastic demands. The NP-hard problem of bank lending decisions was solved by \cite{metawa2017genetic} who obtained with a GA a substantial reduction in screening time, and increase in bank profit.  

\subsubsection{Genetic programming}
The evolutionary approach of the GA inspired a field known as Genetic Programming (GP): evolution-guided design of computer programs to solve a given task, starting from basic operators. How can computers learn to solve problems without being explicitly programmed, how can computers be designed for a given task without being told exactly how to do it? Genetic programming approaches these questions to evolve populations of computer programs, and those have succeeded to evolve correct programs to solve diverse problems, such as optimal control, discovering game strategies, planning, sequence induction, symbolic regression, image compression, robotics... (\cite{koza1992genetic},\cite{koza1995survey}). One of the early applications of this perspective was the use of GAs to engineer cellular automata to perform computations (\cite{das1994genetic}). Recently, related applications have conducted to improve the performance of robots behavior exposed to situations or damage not encountered before (\cite{cully2015robots}), enabling more robust, autonomous robots that evolve similarly to animals. Much progress has been done on the design of real-time adaptive trading systems (\cite{dempster2001real}, \cite{dempster2006automated}, \cite{martinez2009heterogeneous}). 
A very recent review of the genetic programming literature has been presented by \cite{langdon2020genetic}. More recent applications have focused on the design of neural network architectures (\cite{suganuma2017genetic}), production scheduling (\cite{nguyen2017genetic}), classification tasks such as detection of video change (\cite{bianco2017combination}), and the determination of new functions to detect breast cancer (\cite{devarriya2020unbalanced}). In these diverse applications, the GP approach was used for its encoding openness in creating computer programs, and effectiveness in evolving efficient ones. It extends the interest of genetic methods in evolving not only individuals or organisms, but more complex objects such as computer programs. Genetic programming is a research field on its own, but shares so many traits with GAs that we wished to evoke them here. An important feature in this problem solving ability in genetic methods, from GAs to GP, is the ability to generate novelty, and explore new regions in the search space. This has fueled various recent developments, from novelty search to open-endedness and the study of evolutionary systems such as ecosystems, or economic environments.

\subsection{Emergence, novelty and open-endedness}

\subsubsection{GAs are more than optimizers: they create novelty}

The successes of GAs in solving such complex problems have often led researchers to measure their progress with learning curves, fitness curves, and to emphasize the optimizing qualities of GAs. For instance, when training a GA to learn a classification task such as identifying a correct object on a picture, the progress of the population of classifiers is easily measured as the error rate, that decreases with the number of iterations. using a GA to develop a game strategy, an increase in performance factors such as win percentage or number of points scored would be observed. However, it would be restrictive and misleading to only see them as function optimizers (\cite{de1993genetic}). While their problem solving methods have been improved by elitism (\cite{ahn2003elitism}), i.e. keeping the best chromosome alive from one generation to the next, or faster implementations (\cite{deb2002nsga}) the canonical GAs remain deprived of a "killer instinct". They identify quite rapidly promising regions of the search space, but do not locate the exact optimum with a similar speed. Unless the problem is very specialised, in such a way that it is possible to adjust a specialised GA to take profit from this information, the canonical GA is likely to outperform over-exploiting GA-optimizers. Moreover, for such a specialised problem, other traditional methods such as hill climbing, neural networks are usually acknowledged as converging faster to the solution. Altering drastically the way crossover, selection or mutation function in an attempt to turn the GA into a function optimizer may alter its behavior. The resulting biases put on the search process may shrink their desirable robustness and exploration abilities, that are the reasons for which we used the GA as a problem solver in the first place. A key reason for seeing GAs as more than an optimiser is the generation of novelty, which marks a decisive advantage of GAs among other search methods. 

One could interpret the schema theorem as a limitation: for successful evolution to happen, intermediary rewards should exist in the fitness landscape (\cite{de2008evolving}). In other words, the genetic algorithm would need to be rewarded for standing up, if we desired it to learn how to walk and run. This has often led to seeing GAs as vulnerable to \textit{deception}, by the capacity to create a landscape that deceives the GA by hiding a high-fitness region in a larger region of lower fitness (\cite{forrest1991royal}). A related issue has been identified in multi-modal environments (\cite{forrest1991royal}), or functions with several peaks, leading the GA to prematurely converge to one of the optima due to generic drift induced by random fluctuations in the sampling process. The production of novel, diverse solutions is an essential element to tackle these issues. It is a common pattern in GA runs to observe the apparition of mutations that seem to harm fitness in the short term, but open to significant improvements in the long run. GA adaptations with sharing or niching rules to prevent crowding of solutions (\cite{Wirsansky2020hands}), emphasis placed on novelty through novelty search (\cite{lehman2011abandoning}) or quality-diversity search (\cite{cully2017quality}, \cite{pugh2016quality}), provide workarounds for these issues, and draw new avenues for further research. They establish connections with the general problem of artificial intelligence (\cite{clune2019ai}, \cite{stanley2019designing}), open-endedness in evolution (\cite{stanley2017open}, \cite{stanley2019open}) and artificial life.

\subsubsection{Artificial life and emergence}

GAs have been originally seen as promising to model the natural systems that inspired their design. Theories on the behavior and evolution of these systems can be implemented and tested in a GA. They allow to explore variations in details of the theories, simulate phenomena that would be difficult or even impossible to capture and analyse in a set of equations (\cite{mitchell1998introduction}). They also connect evolution with information theory through that computational process lens. They provide by their ability of generate novelty, and to explore large -open..?- search spaces, a useful tool to model the evolution living systems, both natural and human. One of the first of these ecological model was Holland's Echo (\cite{holland1999echoing}), studying how simple interactions between basic agents could foster the emergence of sophisticated higher-level phenomena, such as cooperation, competition, or the rise of communities. These first models of artificial life using GAs were reviewed and commented by \cite{mitchell1994genetic}. Models of artificial, "digital" evolution, are not just simulations of evolution, but mere instances, that are complex, creative and surprising (\cite{lehman2020surprising}). GAs thus carry a great creative potential to model these systems, possibly generating patterns entities or ideas that we would not think about.\par

The evolutionary approach carried by the GA has relevance in the modelling of evolutionary human systems. In economics and finance, the evolution paradigm captures well the inductive theory of learning described by \cite{arthur1994inductive}: agents generalise previous patterns for their future behavior, and adapt through trial and error to a changing environment. The adaptive market hypothesis (\cite{lo2004adaptive}, \cite{lo2019adaptive}) places individual agent evolution in a changing environment with simple heuristic as a driving force for economic and financial dynamics. \cite{palmer1994artificial} created the Santa Fe Institute artificial stock market as an emergent product of evolving agent behaviors. The resulting artificial economy exhibited generated trends, speculative bubbles and crashes. The ecology these evolving agents created exhibited symbiosis, parasitism, arms races, mimicry, niche formation, speciation, all similar to natural ecosystems.  Agents are co-evolving, and adapt their internal models of behavior rules to grow more sophisticated ones, by observing success and failure. The GA was used to govern the evolution of those rules, and model the creation of new ones by exploration, or innovation, providing a plausible mechanism to match empirical patterns (\cite{lebaron1995experiments}). Since then, most of the evolutionary approaches in economics and finance have been carried by agent-based models. Most of the uses of GAs in economics have been applied works: analysis of insolvency risk (\cite{varetto1998genetic}), supply chain design (\cite{altiparmak2009steady}), design of automated trading systems (\cite{dempster2001real}, (\cite{dempster2006automated}, \cite{martinez2009heterogeneous})), prediction of daily foreign exchange rates (\cite{waheeb2019new}) or daily bitcoin prices (\cite{han2019using})) and bank lending decisions (\cite{metawa2017genetic}). There has been very limited theoretical follow up to the pioneering use of GAs by \cite{palmer1994artificial}, as most of the evolutionary economics community shifted largely towards emergent macroeconomics and financial dynamics in multi-agent models. Most of these models including behavior rules encompass a limited form of evolution of these rules, most often restricted to a small number, and with little variation. We argue that in this context, GAs have a role to (re)play. Their ability to generate novelty, explore new strategies, and account for economic learning and innovation, is relevant to domains such as market ecology, that study these economic systems from the participants types or behaviors. Allowing these behaviors to evolve in a larger space, and observing the resulting economic and financial patterns, is a promising and realistic direction for the application of GAs as modelling tools. \\

By using a population-based search method that uses the information from previous evaluations to decide what to evaluate next, GAs achieve a balance between exploration and exploitation that allows them to successfully explore large, rugged search spaces. Their ability to identify global optima in such complex settings has allowed numerous achievements in various fields, from economics and operations research to deep learning and strategical games. GAs have a duality in the sense that they can be used to both optimise a system, and model its evolution towards efficiency. By maintaining diversity, and generating novelty during evolution, they are relevant to study evolutionary systems such as financial markets. These qualities however come with significant challenges and limitations, that the next section presents.

\section{Limitations, challenges and perspectives}
\label{problems}

Genetic algorithms face critical challenges that have limited their applications. This section attempts to identify and explain these main issues, as well as the solutions identified by previous research. First, GAs are criticized for their high computation cost. They are usually computationally expensive to run, that is, they may have to run for a long time to give adequate results, and having a larger population or studying a more complex problem significantly slows them. A second related issue deals with parameter configuration. We exposed in the previous sections the several parameters used by the GA, and how they impact its results. A poorly configured GA may converge very slowly, or even not converge at all. The combined effects of parameters are not trivial, thus the task of configuring a GA has been difficult. Finally, the relevance of the GA is conditional on the relevance of the representation strategy being used. The way entities are represented, the choice of the fitness function, the sampling of the initial population, are critical to having useful and meaningful and satisfying results.

\subsection{Computational efficiency and cost}

Computational efficiency rules the GA convergence speed, and its performance. While they have often been criticized for their computational speed and complexity, and for the constraints efficiency sets on their design, GAs greatly scale with the use of parallelism and adjusted selection methods. 

\subsubsection{Selection methods for computational efficiency}
Evolutionary algorithms have been criticized for the computational complexity, often identified to be of order $O(MN^3)$ where $M$ is the number of objectives and $N$ the population size. Indeed, comparing the respective fitness of $N$ individuals with all others, and selecting up to $N$ individuals for the next generation, contributes to a great computational cost that ought to be reduced as much as possible. In an influential article, \cite{deb2002fast} introduced a faster elitist, multi-objective GA: "NSGA-II". Elitism, i.e. keeping in the next generation the current best chromosome in the population, speeds up the convergence speed of the GA and allows to prevent the "catastrophic forgetting" of good solutions (\cite{rudolph2001evolutionary}, \cite{zitzler2000comparison}). The NSGA-II algorithm (\cite{deb2002nsga}) limited computational complexity to $O(MN^2)$ using domination count, and a sharing rule for draws that gives preference to diversity. This gives us a fast non-dominated sorting procedure and a fast crowded distance operator, while computation is further improved by elitism. This specification is an example of various attempts to tweak evolutionary operators to generate more efficient procedures. \par

Innovations in computation efficiency have also impacted selection methods. In particular, the tournament selection method which operates a rank selection, but over a subset of the population, is computationally more efficient, and more amenable to parallel implementation (\cite{mitchell1998introduction}). Instead of comparing every entity to all the other ones, tournament selection compares it to say, half, or ten, other individuals, to reduce the number of comparisons to operate. The size of the subset required to correctly evaluate dominance of solutions remains however to be determined, and enters in the larger challenge of parameter configuration in GAs, that we analyse in more detail below. Recently, some metrics have been introduced to quantify the complexity and cost of given tasks for a GA: partial evaluation (\cite{rodriguez2020partial}) and fitness landscape analysis (\cite{merz2000fitness}, \cite{pitzer2012comprehensive}, \cite{wang2017population}). While the early applications of GAs looked for improvements in computation efficiency on the operators' side, significant progress has been achieved by \textit{parallelisation}. That is, using computer systems with several processing units (CPUs) or graphical processing units (GPU) to divide the computational work, so that the program runs in parallel. This can allow to reach considerable reductions in computation time. More recently, some attempts have been made to develop interplay between quantum and genetic algorithms, achieving an unprecedented level of parrallelisation. In the quantum approach to genetic algorithms, not only the evaluation of points can be amenable to parrallelisation, but each point itself, due to \textit{superposition}, could take multiple values at the same time. \cite{malossini2008quantum} showed that under this implementation, the computational complexity of a quantum genetic algorithm could be reduced to $O(1)$. When and how quantum computers will become available remains unknown, but promising achievements could be made in this approach, that might reach unprecedented search speed.

\subsubsection{Implicit and explicit parallelism}
Beyond the mostly static setting experimental GAs have focused on, high computational efficiency is necessary to consider more dynamic problems. By the time the problem is solved, it may have changed (\cite{reeves1995genetic}). Adaptation in a sufficiently random, or unstable environment, is very difficult (\cite{mitchell1998introduction}). Natural evolution occurs in natural environments that are also changing over time. However, species evolve relatively much faster than climate, geology or fundamental environment features. If the environment changes faster than the GA populations can adapt, the effectiveness of the search becomes null. GAs have quite rarely been applied in dynamic or unstable environments for this reason. A great challenge for the development and improvement of new GAs resides in the ability to simulate an evolution that is fast enough to cope with the changes in the environment, but diverse and open enough not to overfit one particular environment instance, and stay adaptive. \par
As for any optimization-based procedure, the evaluation function must be fast to compute. However, as the evaluation is repeated a large number of times depending on population size, the number of constraints, and of iterations, it is a concerning issue for GAs (\cite{whitley1994genetic}). Much theoretical effort has been devoted to improve the information acquired from a finite set of evaluation. Alternative paradigms for the determination of the fitness evaluation function have been proposed. In particular, \cite{huang1998genetic} have shown that using a fuzzy fitness evaluation function converged to results that were identical to the ones obtained by a standard GA, while considerably reducing the computation time (\cite{laribi2004combined}). \par

We emphasized the great scaling of GAs with parallelism in the context of their exploration capacities. The great synergy of GAs with parallelism, denoted "implicit parallelism" (\cite{holland1992genetic}) mimicking the massive parallelism at play in natural systems composed of millions of individuals, provides substantial qualities in computation efficiency. GAs are able to test and exploit a large number of locations of the search space by manipulating only a few strings. The first explicit parallel implementations of GAs introduced multi-processor systems, each running a GA on own sub populations, with periodical migrations of the best solutions to other processors (\cite{tanese1987parallel}). This distributed genetic algorithm with migrations was shown to perform better than the traditional one, even when each sub-population was running different parameter settings (\cite{tanese1989distributed}). 

Recently, approaches of parallelism related to data partitioning have been demonstrated as being more efficient in accuracy, efficiency and scalability (\cite{alterkawi2019parallelism}). GAs enhanced with parallelism distributions have performed better than existing algorithms (\cite{tang2017mapping}). With decomposition approaches, both implicit and explicit parallelism are applied. Different sub-populations are being evolved (\cite{chen2018investigating}). Similarly to deep learning and machine learning, GAs have entered in the world of high performance computing, benefiting from the power of GPU architectures. \cite{cheng2019accelerating} provide a comprehensive review of parallelism approaches, and their principal challenges. In the same way GPU computing has transformed deep learning and machine learning, it is yet to percolate in the GA community. When it will, it is likely that such highly parallelisable search technique will take immense benefits from this additional computing power. A final point of computational efficiency and speed, linked with search speed, is co-evolution.

\subsubsection{Co-evolution}
Competition between candidates inspired by the Darwinian survival of the fittest is the drive of the evolution of genotypes operated by the GA. Some authors have extended this framework to not only have competition within a GA, but between GAs. Inspired from co-evolution in biology, various GA applications have designed adversarial GAs to improve optimization in a related fashion to generative adversarial networks. A founding application was done by \cite{hillis1990co}, who found that developing parasite-GAs against a GA trying to perform a classification task, significantly improved the optimization abilities of the latter. \cite{garcia2017investigating} used an adversarial GA network to develop network cyber-defence strategies against attacks. More recently, co-evolution has similarly contributed to applications of GAs in game theory. Recently, the black box optimisation of such adversarial attacks in neural networks has successfully used genetic algorithms (\cite{chen2019poba}). A new state of the art in black box adversarial attacks was thus obtained, evolving more and more successful attacks. In a related fashion, \cite{vie2020blotto} developed two GAs competing to develop optimal strategies in asymmetrical Blotto games. These are simultaneous resource allocation wargames in which two players allocate some resources over different battlefields, wining each battlefield if they have deployed more resources than their opponent. While solving versions of the game with asymmetric resource endowments was not analytically doable, co-evolving GAs allowed to approximate equilibrium strategies in this setting. They developed sophisticated and empirically consistent behaviors such as guerilla warfare, and concentration of competition. Notably, the GA with less resources learned to focus its resources on fewer battlegrounds. Co-evolving GAs achieved a significantly faster convergence than GAs on their own.

\subsection{Parameter configuration}

\subsubsection{Parameter tuning}

The parameter calibration of GAs -population size, mutation rate, choice of operators...- is a critical determinant of its convergence behavior, and computational efficiency (\cite{maulik2000genetic}, \cite{de2020evolutionary}). Parameters can be set at the start and fixed (parameter "tuning"), or changed during search (parameter "control"). A poorly configured GA can prematurely converge to a sub-optimal solution, or not converge at all, or converge so slowly that the entire process is essentially a waste of time. The performance of GAs is a nonlinear function of their parameters. \cite{grefenstette1986optimization} searched for optimal GA parameters using another GA, showing the efficiency of GA as meta-level optimisation techniques. The space of GAs was described as having six dimensions: 
\begin{enumerate}
    \item \textbf{Population size}: a large population favors diversity and mitigates premature convergence, but is detrimental to computation efficiency
    \item \textbf{Crossover rate}: the higher the frequency of crossover, the higher the frequency of introduction of new structures. A too high crossover rate can discard good solutions faster than selection can improve them, while a too low rate may create stagnation with a resulting lower exploration rate.
    \item \textbf{Mutation rate}: a too large mutation rate creates an inefficient random search, while a too low mutation rate fails to prevent a given bit to remain forever in the population or can fail to mitigate premature convergence.
    \item \textbf{Generation gap}: the percentage of the population to be replaced during each generation is optimised with respect to same above the trade-offs.
    \item \textbf{Scaling window}: the reference to which solutions are compared may change their relative fitness, and alter the resulting fitness-proportional selection.
    \item \textbf{Selection strategy}: pure selection, elitist selection, but also other mechanisms, are possible, and induce  particular balance between selection, diversity, efficiency and convergence rate.
\end{enumerate}

As it is simple to understand the impact of one parameter, the others being constant (it has been suggested that most parameters exhibited uni-modal, convex responses (\cite{pushak2018algorithm})), but hard to understand interactions of parameters, \cite{grefenstette1986optimization} developed a meta-level GA to identify high-performance GAs on some numerical test functions. The resulting GAs received a significant boost in performance. Some regularities were identified: mutation rates above 0.05 (or at 0) were usually harmful, and the optimal crossover rate and optimal mutation rates appear negatively correlated. Crossover rate appears to decrease as the population size increases. These insights were crucial in the early applications of GAs and proposed some much used baseline parameter values. \cite{grefenstette1986optimization} and \cite{caruana1988representation} identified general ideal settings for elitist (non-elitist) selection strategies: a population size of 30 (80), a crossover probability of 0.95 (0,45), a mutation probability of 0.01 (0.01). \par

In the spirit of the no free lunch theorem of optimization for which no algorithm works universally for all optimization problems, one needs to tailor the GA to tackle different problems. The practical value of these theoretical results on parameter settings remains unclear: they can be expected to vary depending on problems and the variant operators used by the GA. These early attempts to design optimal GAs remained limited by the set limitations on the space of possible GAs, and the computational cost of the meta-simulations.\\

Recent progress has been made in automatic parameter tuning (automatic algorithm configuration) to eliminate the limitations and drawbacks of manual parameter setting. This process can be described as a meta-optimization process that identifies the set of parameters in a configuration space for a given parametrised algorithm that maximizes a performance metric over a set of problem instances.  \cite{huang2019survey} surveys different state of the art techniques used in automatic parameter tuning, the classification of parameters into numerical (rates) and categorical (operators) types. Different tuning methods do exist: the Simple Generate-Evaluate (GEM) methods generate a set of candidate configurations, evaluate their performance and select the maximizing one Brute-force and F-race (close to dominance selection) approaches are quite popular in SGEMs. Iterative GEMs create new configurations throughout search, and include notably heuristic search-based methods. High-level GEMs use existing tuners and search methods to generate high-quality candidates and compare them. Among these categories of tuning algorithms, Iterative GEMs stand out by their efficiency in exploring the configuration space, and their computation efficiency in doing so, as they use information from previous evaluations to generate new candidates. In particular, heuristic search-based iterative GEMs include iterative F-races (that eliminate candidates as statistic evidence grows against them, adapted to cases of many candidates: see \cite{balaprakash2007improvement}), meta-evolutionary algorithms (meta-EAs) and ParamILS (a versatile local search approach with adaptive capping of runs to avoid unnecessary runs, introduced by \cite{hutter2009paramils}). Two techniques are considered as the current state of the art within meta-EAs. 

The Covariance Matrix Adaptation (CMA)-Evolution Strategies (CMA-ES, \cite{hansen2006cma}, \cite{hansen2016cma}) is a numerical optimization technique that samples candidates according to a multivariate normal distribution. Recombination amounts to changing the distribution mean, and mutations consist in a zero-mean perturbation. The covariance matrix of the variables is updated to maximise the likelihood of occurrence of previously successful solutions, in a related fashion to gradient descent. This method is very fast with small populations, essentially parameter free, and does makes very few assumptions on the underlying objective function. Evolution paths information is taken into account in step-size control to prevent premature convergence and from overshooting the optimal region. \par

The Gender-based GA (GGA, \cite{ansotegui2009gender}) takes advantage of the parallelism and handling of rugged landscapes features of the GA. As the way parameters interact with each others is a priori unknown, the GA becomes a great candidate. They distinguish populations in two categories, and apply different selection pressures. Only the competitive sub-population receives intra-specific competition, and struggle to mate with the noncompetitive sub-population. This part of the population not directly subject to selection allows to diversify the search and prevent premature convergence, while improving the computation cost of the automatic tuner. This work is related to the interesting introduction of genders in the population (\cite{rejeb2000new}, \cite{sanchez2003gendered}) to regulate the mating process in a more natural way, outperforming the GA without genders in graph partitioning problems in solution quality and the number of generations required to attain the solution. As \cite{huang2019survey}, the tuning process being computationally expensive could be improved by reducing the configurations to evaluate, or reducing the average cost of evaluation. Improving the computation efficiency of these parameter tuning methods hence remains an area of future investigation. Though capable of reaching the global optimum, and to return good solutions at any point of the search, meta-EAs do require a large number of evaluations, which entails computation cost issues with large parameter spaces, and struggle in handling categorical parameters (\cite{huang2019survey}, \cite{de2020systematic}). However, the GPU could spark a revolution in making these computation-heavy parameter configuration search methods feasible and more precise by searching over a large space of configurations. \par

In the larger field of algorithm automatic configuration, evolutionary algorithms are both subject and answer to the challenge of parameter tuning. These works on parameter tuning have been recently been complemented with an increasing attention on parameter control, where the values of the parameters are changing during the run according to some deterministic, adaptive or self-adaptive strategies. 

\subsubsection{Parameter control}

A single run of the GA is a stochastic path, intrinsically adaptive. Sticking to rigid parameters is in contrast with its spirit (\cite{grefenstette1986optimization}). It is intuitive that different parameters settings may be optimal at different periods in time (\cite{back1992interaction}, \cite{forrest1993optimal}). Large mutations could be explored in the early steps to maximize exploration of promising regions of the search space, increasing selection pressure in later stages to narrow down the search to the optimal chromosomes, as does Boltzman selection (\cite{chen2019improved}). Allowing a GA to dynamically modify its own parameters during a run was first suggested by \cite{grefenstette1986optimization}, noting however that the evaluations done in some time interval would probably not be sufficient to conduct a realistic assessment of the search traits performance. Benefits and early approaches were surveyed by \cite{eiben1999parameter}. \par

In the deterministic parameter control paradigm, parameters are changed by a fixed, predetermined rule, which can improve GA performance, but what the optimal rule depends again on the context. Adaptive parameter control uses some feedback from the search to determine the changes to implement in the parameters. However, the updating mechanism used to control parameters is externally set. Both these methods suffer from the difficulty of identifying such optimal rules of change (\cite{eiben1999parameter}), an issue addressed by the use of self-adaptive parameter control. \\

Natural evolution is itself a powerful meta-learning algorithm (\cite{stanley2019designing}), and was seen quite early as a natural development for GAs, with first positive theoretical results appearing with \cite{greenwood2001convergence}. \cite{grefenstette1986optimization} presented self adaptation as two dual searches taking place, one search for a solution to a given problem in a search space of possible candidates, and a second for an optimal algorithm setting in the space of possible algorithms. Adapting only the mutation rate for example, would restrict this second search to the space of algorithms with fixed mutation rate. Population size itself in natural systems is controlled by complex ecological interactions (\cite{mitchell1998introduction}). \par

A line of research has explored self adaptation of operators rates, to let the GA choose the probability of distributions of operators. \cite{davis1989adapting} identified a key challenge for self adaptation of GAs to work: having the GA adaptation rate -and find ways to measure it- match the population adaptation rate. That "synchronization" problem appears to be the obstacle towards efficient self-adapting GAs. \cite{hassanat2017enhancing} enhanced GA performance with multiple crossover operators, and a meta-selection of the best to use. 
Meta-GAs assembled in a population with migrations were found appropriate to find good generalist parameter configurations, and to achieve performance close to specialist configurations for given problems (\cite{clune2005investigations}). Their results also outlined the existence of shifts on the optimal parameters or operators through time, notably from crossover types, which was beneficial in short term performance, but detrimental in the longer run. \cite{clune2005investigations} also show how the ability to adapt exhibited by meta-GAs could lead to premature convergence to local optima. It is possible that those results apply in a specific context with a relatively low population (36 different GAs), or some restrictions of the search space, but they outline that while the meta-GA approach has some potential, it still faces important issues that need to be addressed to achieve it, notably the chain issue of finding the parameter configuration of higher level meta-GAs. In a related fashion, natural evolution failed to optimise mutation rates on rugged fitness landscapes, and selects sub-optimal mutation rates for their short term advantages (\cite{clune2008natural}). It appears that the dual search of a GA in the space of GAs, and the solution to a problem in a solution search space, is more complex, and amplifies the challenges they face. Parameter control is undoubtedly non-universal, and specific to each GA run (\cite{huang2019survey}). We highlight in the next section some promising directions for fundamental breakthroughs that may help us come back with new tools to these topics of parameter control by self-adaptive algorithms. Very recently, \cite{case2020self} and \cite{dang2016self} demonstrated theoretically and experimentally that self adaptation of GAs where parameters such a the mutation rate, are encoded with the individual chromosomes, leads to significant speedups, achieve optimal parameters as if they were known in advance. Evolutionary algorithms enhanced with self adaptation were found to have an asymptotic speed improvement over the state of the art solution for their considered problem (\cite{case2020self}). 

\subsection{Realism and robustness}

It is not enough for GAs to be computationally efficient, and well parametrised. We also desire them to be robust in their trajectories, open in their exploration capacities, and realistic in their design. Choices of representation of individuals, and design of the fitness landscape, are essential choices to ensure realism and robustness of the approach. We open towards a greater embrace of evolution, to not only evolve solutions in complex environments, but also the GA itself, and address these limitations all together.

\subsubsection{Initial population sampling}

This concern for diversity starts from the very initialisation of the GA. Bias in the first generation is likely to induce further bias in search, leading potentially to premature convergence and local optima. Sampling error was early identified as one of the main difficulties faced by GAs (\cite{forrest1991royal}), and can induce premature convergence (\cite{forrest1993makes}). In matter of instance of evolution, studying whether and how initial conditions impact the GA outcomes provides as well valuable information on the evolution process. However, for more practical purposes, some more rigor on the initial sampling can greatly benefit GA applications. In high dimensional spaces, new methods to generate initial populations allow to maintain diversity with limited population sizes. While little attention has been paid on establishing formal desirable conditions for this initial diversity, significant results for searching in very large spaces have been achieved starting from "unbiased" initial populations (\cite{deaven1995molecular}). This unbiasedness has often been achieved by simple random uniform sampling. As the object being considered by the GA grows in dimensions and complexity, random uniform sampling requires a substantially higher population size to satisfy this unbiasedness objective. To maintain the diversity of the GA initial population without sacrificing computation efficiency with excessively large population sizes, we may be interested in alternative initial seeding techniques. Recent advances have suggested advanced sampling methods as a promising means to achieve better robustness and diversity in high-dimensional simulation exploration. Latin Hypercube Sampling (\cite{helton2003latin}, see application and discussion by \cite{jing2019rbf}) and Sobol Sequences (\cite{sobol1967distribution}), that consist in generating low discrepancy sequences, i.e. quasi uniform sets of points in a high dimensional space, are now being used in the analysis of simulation models (\cite{Reuillon2013}), and can similarly improve the starting points of the GA in the search space. While the current practice of GAs is focused on random initial samples (\cite{yakovlev2019optimization}), we maintain that the danger of sampling bias for GAs optimisation is significant, increasing in higher dimensions, and can be mitigated by the above sampling methods. 

\subsubsection{Representation}

One of the most crucial limitation of GAs is representation, most often denoted encoding, that denotes how the action space is modelled. Various encoding techniques exist in the literature. The most popular, the binary encoding, uses strings composed of 0s and 1s. It has the advantage of containing a higher degree of implicit parallelism, as a GA instance will contain more schemas, and most of the mutation, crossover operators have been built around the binary encoding. The encoding representation commonly used in GAs admits many variants, and exhibits a high versatility. However, binary encoding might be too narrow an encoding for some problems. Some may require integer values, some may contain strings or operators. In other cases, the number of bits necessary to encode a certain space of solutions in a binary way would simply be too large for this encoding to function. The encoding representation indirectly constrains the search space, with the issue of partial cover. Only a robust, careful representation design allows to cover its fully diversity (\cite{juzonis2012specialized}).  A founding analysis on encoding in Genetic algorithms has been done by \cite{ronald1997robust}. He notably defined the nine desirable properties for a robust encoding in the context of the schema approach:
\begin{enumerate}[noitemsep]
    \item Embodies the problem-relevant fundamental building blocks
    \item Is amenable to a set of genetic operators to operate selection, crossover and mutation
    \item Minimises \textit{epistatis} (expression of one gene suppresses the action of others)
    \item Allows a tractable mapping to the phenotype that allows fitness to be measured
    \item Exploits an appropriate mapping from the genotype to the phenotype
    \item Embodies feasible solutions and discards illegal candidates
    \item Suppresses isomorphism and mitigates redundancy: many genotypes converging to a same solution point
    \item Uses the smallest cardinally of an alphabet for the gene values, the binary being the best if relevant
    \item Represents the problem at the correct level of abstraction
\end{enumerate}

To avoid \textit{genetic hitchhiking}, in which some low fitness sequences associated with highly successful alleles are maintained in spite of their low overall performance, and reduce the size of the representations GA have to search, \cite{schraudolph1992dynamic} proposed a dynamic parameter encoding mechanism that adjusts in size as evolution occurs. \cite{kumar2013encoding} surveys the different encoding schemes: binary, octal, hexadecimal, permutation, value, tree encoding, and restricts their design to the principle of minimal alphabet, and principle of meaningful building blocks.\\

The use of GAs to evolve neural network architectures inspired a lot of research on encoding and representation. As the network size grew, so did the size of the required chromosome to encode the network structure, which led to significant issues in performance (how high a fitness can be obtained) and efficiency (how long does it take to obtain this high fitness result). Direct encoding methods represent each network connection, and struggled in encoding repeated structure, such as symmetry in a network. For complex networks, encoding the network adjacency matrix in the chromosomes may become huge and intractable for any search algorithm. A set of 100 nodes would require chromosomes of 5000 bits. A set of 1000 nodes would have to be encoded in 500,000-bit strings to represent each connection. One solution identified by \cite{kitano1990designing} was grammatical encoding: encode network as grammars of structure operators, in a related fashion to genetic programming. This representation requires shorter chromosomes, as the GA evolves sets of building instructions, rather than the network structure itself. Grammatical encoding hence mitigates that issue of complexity, and can inspire innovations in encoding representations, in order to design more robust or efficient networks in neural nets, security, communications, production, supply-chain, or financial contexts. \par

Indirect encoding has enjoyed a recent popularity (\cite{stanley2019designing}): the genotype is a formula, a set of instructions for generating the network, rather than a direct edge by edge encoding. Some approaches have been based on Cartesian genetic programming where the encoding being evolved is machine code with some operators (\cite{khan2013fast}, \cite{turner2014neuroevolution}). Inspired from genetic regulatory networks, \cite{mattiussi2007analog} have developed the Analog genetic encoding, that allows complexification and decomplexification of the network during the evolutionary process. How can 100 trillion connections and 100 billion neurons be encoded in a DNA encoding of 30,000 genes? (\cite{stanley2019designing}) Indirect encoding aims at using regularity such as symmetries or motifs, to improve compression. This has been connected with canalization: capacity of natural indirect encoding to yield robust, adaptable evolutionary paths of development (\cite{le2013evolution}).\\

As \cite{grefenstette1986optimization} point out, these representations, whether binary, finite alphabet in fixed-length chromosomes, is intrinsically inconsistent with the adaptive, evolution framework of the GA. These encoding methods are not adaptive, and as such, they restrict the search space, potentially limiting it to sub-optimal choices of lesser complexity. Any fixed length representation limits the complexity of the candidate solutions, hence the search space (\cite{mitchell1998introduction}). It could even be argued that evolution not only makes the genotypes bigger in size, more complex and more adaptive, but could change the genetic encoding and the mapping between genotypes and phenotypes themselves (\cite{mitchell1998introduction}). As we consider the chromosomes as carrying evolving information, but also as evolving objects, we walk even further into an evolutionary paradigm. As we seek further inspiration from biology and genetics to understand how nature is encoded and mapped to phenotypes, we may move from manual design of representation to evolved representations, in which the design principles we enumerated are the product of evolution, rather than constraints on manual design. Embracing the evolution paradigm to a larger extent leads GAs to open-endedness, and vast uncertainty on their resulting behavior or properties, that is largely left to analyse, but that may contain essential innovations for the practice of GAs.

\subsubsection{Fitness evaluation}

While the evaluation function is often contained in the problem formulation, its design can be a difficult task (\cite{whitley1994genetic}). Digital evolution is surprisingly creative in exploiting misspecified fitness functions, create unintended debugging of simulated environments (\cite{lehman2020surprising}). The design of the fitness function may reflect some preconceptions of the experimenter on the form of the solution, include unintended loopholes that are easy for evolution programs to exploit, in the same ways well-intentioned metrics in human societies can have detrimental effects by the pressure to optimise them. Well-intentioned quantitative measures of fitness can be maximized in counter-intuitive, glitchy ways. In some other cases, digital evolution programs learned unintended regularities instead of learning the complex behaviors experimenters were aiming for (\cite{ellefsen2014neural}), or took advantage of simulation bugs or design flaws that were previously unseen. Various stories of digital evolution algorithms "outsmarting" their creators are compiled by \cite{lehman2020surprising}. \par

Evidently, fitness is not exogenous in nature, and there is no such physical thing as an objective fitness evaluation function. At best, this function would appear rather related to fuzzy logic (\cite{huang1998genetic}). While an objective function is certainly relevant for optimization problems, subject to the design issues we highlighted above, modelling more complex, evolutionary systems could benefit from endogenous fitness determination (\cite{holland1999echoing}). The fitness, or qualities of a given phenotype, may heavily depend not only on a fixed environment, but also on the other phenotypes present in the environment. When those instabilities in the environment are well understood, they can be included in the fitness evaluation function (\cite{vie2020guessing}). The GA evolving under this more uncertain environment was able to both identify the two Nash equilibria of the variants, and to develop optimal mixed responses in between. \par

Evolving intelligent agents requires them to explore different selves, alternative ways to represent them, and to modify them. One possible direction could be to develop a co-evolutionary paradigm in which the fitness landscape evolves with the agents. . In fact, endogenous fitness opens to multiple phenomena observed in natural systems: predator-prey relationships, symbiosis, crowding effects. Interactions within the GA allow emergence of an evolving, autonomous ecology of phenotypes (\cite{smith2000emergence}), leading to both optimization, and understanding of the ecology behavior. It connects genetic algorithms with artificial life (\cite{mitchell1998introduction}). Endogenous, not explicit, fitness, is at the root of open-ended, creative, surprising digital evolution (\cite{lehman2011abandoning}, \cite{lehman2020surprising}). These directions are likely to require progress in parameter configuration, computation efficiency to become possible, and will most likely create new challenges on how to understand such complex systems, but carry promising insights.

\section{Conclusion}
\label{conclusion}


In this review, we have attempted to present in a condensed way the different properties of genetic algorithms, their merits as exploration and optimisation heuristics, and the challenges they face, notably in computation efficiency, parameter configuration, realism and robustness. We hope to have well covered the search space in this objective and to have identified its most promising points of its landscape. \par

Genetic algorithms are particularly suitable to solving sparse problems in large, rugged search spaces. They require very few assumptions on the fitness landscape properties, deal well with existence of local optima, or multiplicity of extrema. They are applicable to a wide range of problems, and are quite efficient in returning good solutions quickly. They have achieved particular performance in the evolution of neural networks, multi-objective optimisation and genetic programming. They are more than optimisers, as they generate novelty, are directly connected to artificial life as they foster emergence of complex digital ecologies, and a clear relation with open-ended evolution that can greatly improve the modelling of evolutionary, adaptive systems. Genetic algorithms have been facing significant issues in computation efficiency and cost, and in parameter configuration. Improving the efficiency of evolution as it is digitally encoded will undoubtedly involve parallelism, and the evolution of parameter configurations themselves through self-adaptation. Advances in initial sampling methods can overcome sampling bias issues.\\

Further inspiration from biology to incorporate in a computational form some ecological and genetic interactions, will undoubtedly allow practitioners to develop more sophisticated evolutionary algorithms, capable of evolving in changing, or more complex environments. Phenomena of gene duplication, translocation, dominance, sexual differentiation, regulatory networks, have just emerged in the field. Epistasis -interactions between mutations- and pleiotropy -mutations at one locus affect several traits- are crucial in complex and realistic fitness landscapes. New developments on "Structural Genetic Algorithms" (\cite{vie2021sga}) are attempting to incorporate these key biological phenomena to allow GAs to evolve more complex and more adaptive objects, expanding their capabilities. Further connections with mathematical genetics and statistical mechanics are likely to provide new theoretical grounds to understand the behavior of genetic algorithms, and the properties of their operators. The technological progress in computing power and the GPU revolution (\cite{cheng2019accelerating}) has pushed deep learning to the front of machine learning research, and exciting times are ahead as GAs will benefit from it, with their natural scaling with parallelism (\cite{stanley2019designing}). \par

The presence of a tremendous diversity of organisms stemming from natural evolution, arguably involves the evolution of evolvability itself (\cite{huizinga2018emergence}). \cite{lehman2018safe} proposed a safe mutation approach that applies mutations through output gradients, to improve the benefits from mutations, but evolution of mutations towards beneficial ones may also be the product of the evolution of evolvability and pleiotropy, rather than a manual addition. The ability to select some dimensions of variation to be more or less likely to be explored through mutation, a phenomenon known as developmental canalization, is barely emerging in computational simulations of evolution, but could allow the field of GA to tackle more important challenges. Open-ended, divergent evolutionary processes may be necessary for attaining this evolution of evolvability (\cite{huizinga2018emergence}). \par

The latest developments of genetic algorithms towards meta-learning architectures, learning how to learn, and the endogenous generation of learning environments, have placed AI-generating algorithms as a credible means to produce general AI (\cite{clune2019ai}). Open-endedness in their evolution could lead them to produce interesting and increasingly complex discoveries, and so indefinitely (\cite{stanley2019designing}). The first creator of GAs, John Holland, noted that such computer programs that evolved in ways similar to natural selection, could solve complex problems even their creators do not fully understand. We add that they will do so and generate surprising novelty, provided they are endowed with a representation, a landscape that allows to go beyond manual design, and places all GA components, parameters to encoding, subject to open-ended evolution.

\subsection*{Acknowledgments}

We wish to particularly thank Jeff Clune, Rama Cont, Melanie Mitchell, Maarten Scholl, Lisa Soros for their previous time and remarks.

\subsection*{Funding}

This publication is based on work supported [or partially supported] by the EPSRC Centre for Doctoral Training in Mathematics of Random Systems: Analysis, Modelling and Simulation (EP/S023925/1)








\printbibliography
\end{document}